\documentclass[runningheads]{llncs}

\usepackage{accv}

\usepackage{accvabbrv}

\usepackage{graphicx}
\usepackage{booktabs}
\usepackage{amsmath}
\usepackage{amssymb}
\usepackage{multirow}
\usepackage{xcolor}
\usepackage{color,soul}
\usepackage{tikz}
\usepackage{import}
\usepackage{relsize}
\usepackage{fontawesome5}

\usetikzlibrary{arrows.meta, positioning, shapes.geometric, shapes.symbols, fit, backgrounds, calc}
\newcommand{\modelname}{O-DINOv3}

\usepackage[accsupp]{axessibility}  

\usepackage[pagebackref,breaklinks,colorlinks,citecolor=accvblue]{hyperref}

\usepackage{orcidlink}

\begin{document}

\title{Align or Fail: The Hidden Gap in Cross-Domain Panoptic Segmentation} 


\author{Huy Hoang Nguyen\inst{1}\orcidlink{0009-0004-9033-7068} \and
Oliver Zendel\inst{1}\orcidlink{0000-0001-8097-1226} \and
Tobias Glück\inst{1}\orcidlink{0000-0003-1497-6138} \and
Minh Nhat Vu\inst{2}\orcidlink{0000-0003-0692-8830} \and
Andreas Kugi\inst{1,3}\orcidlink{0000-0001-7995-1690}}

\authorrunning{H. H. Nguyen et al.}

\institute{AIT Austrian Institute of Technology, Austria 
\and
VinUniversity\\
\and
Automation and Control Institute (ACIN), Technical University of Vienna\\}

\maketitle

\begin{abstract}
Vision foundation backbones are often treated as drop-in upgrades for improving region proposals in open-vocabulary panoptic segmentation (OVPS). Yet, improvements in backbones often fail to translate well into downstream cross-domain performance.
Using ODISE, a diffusion-based OVPS model with a CLIP-based recognition branch, as a testbed, we replace its diffusion visual extractor with DINOv2, ConvNeXt, or DINOv3, and find that backbone upgrades can introduce a grouping-recognition mismatch: an alignment gap between dense-region features and CLIP text embeddings.
Across source and near-domain, the newer DINOv3 variant improves source-domain performance but collapses on several cross-domain targets, whereas other baselines, interestingly, show the opposite. Stage-wise diagnostics, linear probes, and region-text alignment analyses indicate that DINOv3 embeddings retain semantic information but are geometrically incompatible with CLIP.
We test adaptations, including text-side adapters, conformal mask retention (CMR), and a lean 2.7M-parameter visual-side LoRA, and find that text-side adaptation is more sensitive to vocabulary overlap. CMR provides only limited and target-specific gains, indicating that retention alone cannot resolve the full cross-domain failure. To understand why ADE20K resists improvement across all interventions, we evaluate OVSeg, CAT-Seg, and MaskCLIP. This separates two failure axes: a recoverable visual-to-CLIP interface gap that visual-side LoRA largely repairs, and a vocabulary-boundary limitation shared by the evaluated CLIP-based recognition systems that requires different remedies.

\keywords{open-vocabulary panoptic segmentation \and cross-domain transfer \and vision-language alignment}
\end{abstract}

\section{Introduction}
\label{intro}

\begin{figure*}[t]
\centering
\resizebox{\textwidth}{!}{
    \input{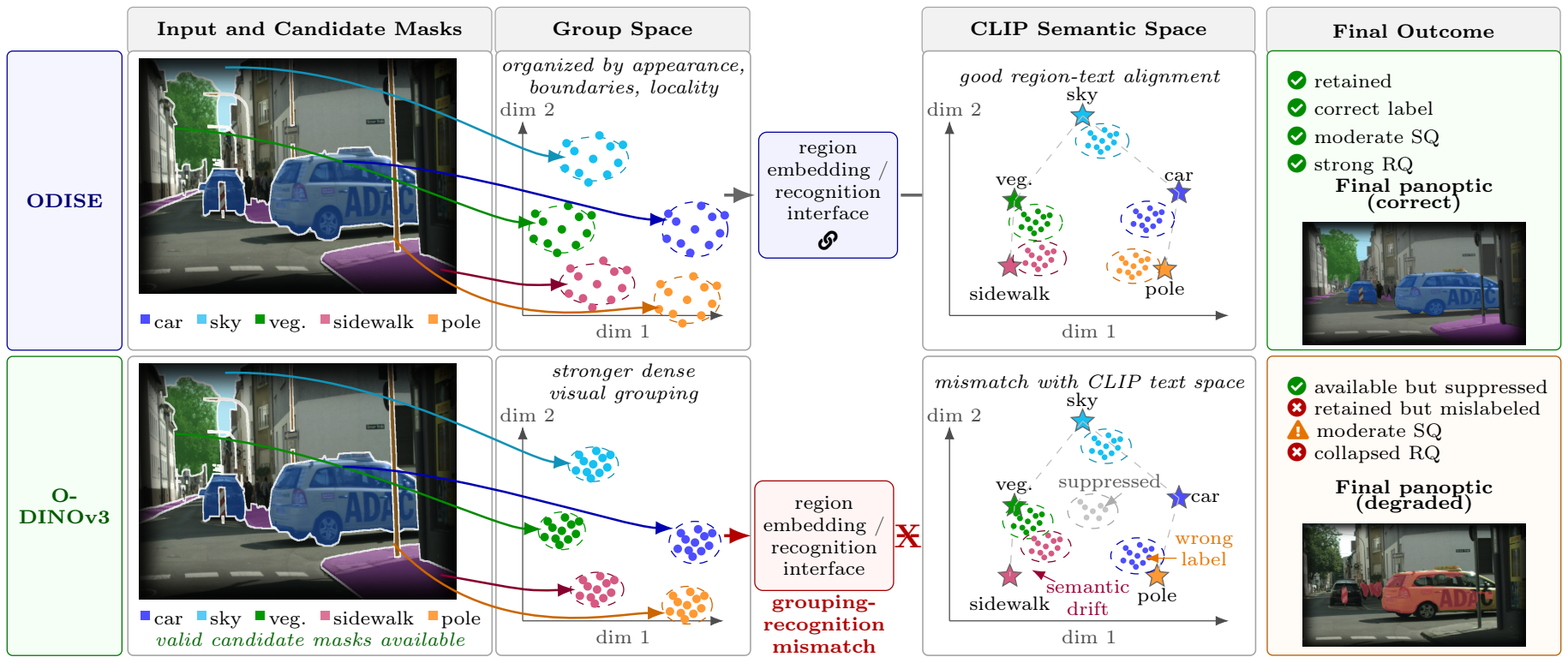}
}
\caption{
\textbf{Grouping-recognition mismatch in cross-domain OVPS.}
A visual backbone may produce valid candidate masks, but the frozen CLIP recognition head can suppress or misclassify them when region embeddings are poorly aligned with text prototypes.
}
\vspace{-1em}
\label{fig:concept_grouping_recognition}
\end{figure*}

Open-vocabulary panoptic segmentation (OVPS) extends panoptic segmentation~\cite{kirillov2019panoptic} beyond a fixed training taxonomy. Given an image, the model must jointly partition the scene into non-overlapping thing and stuff regions while assigning each region a label from a target vocabulary that may differ from the categories seen during training~\cite{xu2023open,yu2023convolutions}. This setting is increasingly important for real-world perception, where deployment domains routinely differ from curated source datasets in visual appearance, annotation policy, and category vocabulary~\cite{wu2024towards}.

Most recent OVPS systems follow a mask-and-recognize design: a segmentation module proposes candidate regions, and a vision-language recognition branch scores each region against text embeddings from the target vocabulary~\cite{kirillov2023segment,vs2024possam,yu2023convolutions}. Improving this pipeline by swapping in a stronger visual backbone seems straightforward, as improved visual grouping should yield better proposals. But this reasoning overlooks a structural coupling that runs through the whole pipeline: a candidate region must not only overlap a ground-truth segment, it must also survive inference-time filtering and produce an embedding that the frozen CLIP-based recognition branch can read reliably.

This coupling makes backbone scaling non-trivial. Dense self-supervised foundations such as DINOv2~\cite{oquab2023dinov2} and DINOv3~\cite{simeoni2025dinov3} learn features well-suited to correspondence, part structure, and spatial grouping. But their geometry emerges from visual self-supervision, whereas CLIP's joint embedding space is shaped by image-text contrastive learning. Whether a backbone that excels at visual grouping remains compatible with a frozen CLIP readout under domain shift is not obvious. Hence, we study this through a controlled intervention on ODISE~\cite{xu2023open}. ODISE is a useful testbed because it separates visual extraction, mask decoding, CLIP recognition, and panoptic inference into distinct components. Replacing the diffusion visual extractor while keeping everything else fixed lets us isolate how changes in visual representation propagate to cross-domain OVPS performance.

We hypothesize that this is a \emph{grouping-recognition mismatch}: DINOv3 can provide strong visual grouping while producing region embeddings that the frozen CLIP readout cannot reliably classify, thereby suppressing or mislabeling otherwise useful proposals. Figure~\ref{fig:concept_grouping_recognition} illustrates this hypothesis.
We investigate it using stage-wise diagnostics across in-domain COCO~\cite{lin2014microsoft}, near-domain Pascal~\cite{everingham2010pascal,mottaghi2014role}, and five cross-domain benchmarks: (Cityscapes~\cite{cordts2016cityscapes}, Mapillary~\cite{neuhold2017mapillary}, WildDash2~\cite{zendel2022wilddash2} (WD2), RailSem19~\cite{zendel2019railsem19}, and ADE20K~\cite{zhou2017scene}). A linear probe on the same region embeddings recovers substantial classification accuracy beyond the frozen CLIP readout, indicating that semantic information is retained but not readily accessible through the CLIP recognition interface. We then intervene at different points in the pipeline. Text-side adaptation substantially recovers performance on urban and rail targets but degrades on ADE20K, whereas a capacity-matched visual-side Low-Rank Adaptation (LoRA)~\cite{hu2022lora} provides more consistent cross-domain recovery. This comparison indicates that adaptation placement matters beyond parameter count. Finally, post-hoc threshold calibration via conformal mask retention (CMR)—which statistically adjusts the mask survival threshold based on top-class probability margins—provides only limited additional gains, suggesting that proposal suppression is not the dominant bottleneck once the visual representation is adapted.

ADE20K presents a different pattern: neither text-side nor visual-side adaptation improves performance, while the unadapted O-DINOv3 model remains close to ODISE. To investigate whether this is a DINOv3-specific problem or a more fundamental limitation, we evaluate three open-vocabulary segmentation systems (OVSeg~\cite{liang2023ovseg}, CAT-Seg~\cite{cho2024catseg}, MaskCLIP~\cite{tu2023open}) on ADE20K with the same COCO-overlap/non-overlap class split. All three exhibit a substantial performance gap between COCO-overlap and non-overlap categories. This consistent pattern across different CLIP-based segmentation systems suggests a broader vocabulary-boundary limitation that the DINOv3–CLIP interface alone cannot explain.

Our main contributions are:
\begin{itemize}
    \item \textbf{Diagnostic framework and bottleneck identification.} We introduce a cascade diagnostic framework alongside a backbone substitution study. This reveals that while stronger visual features improve source-domain OVPS, cross-domain collapse is largely attributable to a mismatch at the visual-to-CLIP interface rather than proposal generation alone.

    \item \textbf{Alignment optimization and systemic limits.} Through capacity-controlled comparisons, we demonstrate that visual-side LoRA outperforms a capacity-matched Large Text Adapter (LTA) on the evaluated urban and rail targets. This shows that the placement of adaptation matters beyond parameter count for achieving cross-modal alignment. Moreover, we show that the vocabulary-boundary performance drop (e.g., the ADE20K non-overlap gap) is consistently observed across multiple CLIP-based methods and architectures.
\end{itemize}

\section{Related Work}
\label{relatedwork}

\subsection{Open-Vocabulary Segmentation and Diagnostic Analyses}
\label{rela:ovps}

Representative mask-and-recognize systems such as ODISE~\cite{xu2023open} and FC-CLIP~\cite{yu2023convolutions} combine mask generation with CLIP-based recognition; other systems improve these pipelines through CLIP adaptation, language-aware decoders, cost-volume aggregation, and unified segmentation training~\cite{xu2023san,cho2024catseg,zhang2023openseed,zou2023xdecoder,qin2023freeseg}. Diffusion generative features have also been shown to be effective region descriptors: Seg4diff~\cite{kim2025seg4diff} demonstrates that internal features of text-to-image diffusion transformers encode visual-semantic structure sufficient for open-vocabulary recognition.

Several studies analyze where OVPS pipelines lose valid predictions. Oracle analyses~\cite{vsaric2025holds} show that valid masks may be discarded through no-object filtering or recognition failures, and recent methods such as OVRCOAT~\cite{ovrcoat} explicitly mitigate objectness bias and region-to-text misalignment to improve mask retention and classification in in-domain OVPS. Other work similarly finds that mask classification can dominate the error budget, motivating stronger region-text alignment~\cite{huang2025enhancing,chng2024adapting,li2025mask}. Training-free methods such as ReME~\cite{xuan2025reme} take a complementary data-centric route, expanding open-vocabulary coverage through retrieval-augmented category prototypes rather than learned adaptation. Rather than redesigning the recognition module, we fix the downstream OVPS pipeline and use backbone substitution as a diagnostic perturbation, tracing how the change propagates through proposal availability, retention, and label accuracy. We further compare whether the resulting interface gap is more effectively repaired on the text side or on the visual side.

\subsection{Foundation Models, Visual-Semantic Alignment, and Domain Shift}
\label{rela:dense-vfm}

Self-supervised vision foundation models such as DINOv2 and DINOv3 learn dense features useful for correspondence, grouping, and spatial structure but may not remain aligned with the CLIP-based recognition interface in OVPS pipelines. Methods such as RegionCLIP, OVSeg, and MaskCLIP explicitly bridge dense visual features and text-aligned recognition through region-level contrastive learning or mask-adapted CLIP features~\cite{zhong2022regionclip,liang2023ovseg,tu2023open,jiao2024collaborative}. CLIP-Adapted~\cite{ge2025clipadapted} extends this line by mapping dense visual features into recognition-compatible text space through structured region-text supervision. We use LoRA~\cite{hu2022lora} to test whether the resulting interface gap is recoverable: a small low-rank update to DINOv3 region features trained only on the source domain can restore cross-domain transfer if the failure is primarily due to misalignment with the recognition interface.

Cross-domain OVPS is additionally challenging because target datasets differ in image statistics, category granularity, and annotation policy. ADE20K includes many categories outside the COCO taxonomy; urban and rail datasets introduce different visual conventions. Recent domain-generalization work uses augmentation, contrastive learning, and vision-language alignment to improve cross-domain robustness~\cite{schwonberg2025domain,benigmim2024collaborating,wei2024stronger}. Our contribution is diagnostic: we separate which failures are recoverable by visual-side adaptation and which persist across the evaluated CLIP-based recognition systems at the vocabulary boundary.

\section{Methodology}
\label{methodology}

\subsection{Problem Formulation and Hypothesis}
\label{metho:problem}

\paragraph{Task definition.}
Given an image $I$ and a target vocabulary $\mathcal{C}_{\mathrm{target}}$, open-vocabulary panoptic segmentation (OVPS) predicts non-overlapping segments $\{(M_i,c_i,s_i)\}$, each with a binary mask $M_i$, a category $c_i \in \mathcal{C}_{\mathrm{target}}$, and a confidence $s_i$. Results are evaluated with panoptic quality $\mathrm{PQ} = \mathrm{SQ} \times \mathrm{RQ}$~\cite{kirillov2019panoptic}, where SQ measures mask overlap and RQ measures detection and labeling correctness.

\paragraph{Grouping-recognition mismatch.}
For a candidate mask to improve OVPS, it must satisfy two conditions: it should overlap a ground-truth region, and its region embedding should be compatible with the correct CLIP text prototype. We define a \emph{grouping-recognition mismatch} as the failure of the second condition despite satisfying the first. In this case, visually plausible regions are available but may be incorrectly scored or mislabeled by the recognition interface. We diagnose this failure through SQ/RQ decomposition, stage-wise diagnostics, region-text alignment, and lightweight adaptation.

\subsection{Controlled Backbone Intervention}
\label{metho:controlled}

We use ODISE as a modular testbed because it separates visual feature extraction, mask decoding, CLIP-based recognition, and panoptic inference. This allows us to change the visual representation while keeping the recognition and inference interface fixed. Our main intervention replaces ODISE's diffusion-based feature extractor with DINOv3 and denotes the resulting model as \modelname{} (retaining the 'O' from ODISE). To determine whether observed degradation is specific to DINOv3 rather than to backbone replacement in general, we also evaluate ODISE-based DINOv2 (O-DINOv2) and CLIP-aligned ConvNeXt (O-ConvNext) controls under the same downstream pipeline.

We additionally rerun the released FC-CLIP checkpoint as an external reference. FC-CLIP is not a controlled backbone intervention in our study, but it contextualizes the ODISE-based results by providing an external reference on the same target datasets.

\subsection{Stage-Wise and Alignment Diagnostics}
\label{metho:stagewise}
\label{metho:regiontext}

We decompose each ground-truth segment into three stages: \textbf{candidate availability} ($\mathrm{IoU}(m,g)>0.5$ for mask $m$ and ground-truth segment $g$), \textbf{proposal retention} (the proposal survives no-object filtering, objectness scoring, and panoptic fusion), and \textbf{semantic assignment} (the retained proposal receives the correct label). This cascade separates missing masks, suppressed valid masks, and wrong-label errors that final PQ conflates.

For each matched proposal with ground-truth class $y$, we measure region-text top-1 accuracy, correct-class similarity, and alignment margin, all computed with the same projected embeddings used for OVPS recognition. A lightweight linear probe on the same embeddings tests whether DINOv3 region features contain class-discriminative information that the frozen CLIP branch does not use reliably. We apply COCO-overlap and non-overlap splits on ADE20K to distinguish interface-related failures vs. difficulties associated with the vocabulary boundary.

\subsection{Interventional Probes and External Benchmark}
\label{metho:interventional}
\label{metho:external}

We use four lightweight probes to test where the recoverable failure lives and whether adaptation placement matters beyond parameter count. \textbf{Small text adapter} (+0.10M): two-layer MLP on CLIP text embeddings. \textbf{Large text adapter (LTA)} (+2.62M): same design, 1024-dimensional bottleneck, capacity-matched to LoRA r16 (+2.7M). \textbf{Visual-side LoRA}: rank-16 LoRA on the last 12 DINOv3 blocks. \textbf{Retention calibration (CMR)}: Inspired by conformal prediction~\cite{angelopoulos2023conformal}, Conformal Mask Retention (CMR) addresses proposal suppression without retraining weights. In standard OVPS inference, candidate masks are discarded if their combined objectness and classification confidence falls below a fixed heuristic threshold. CMR instead computes a non-conformity score from the probability margin between the top-1 and top-2 predicted classes on a held-out source calibration split, statistically lowering and calibrating the survival threshold so that valid but low-confidence cross-domain proposals are retained. LoRA+CMR tests whether retention calibration adds gain once the representation is adapted. By placing LTA and LoRA at the same parameter budget, we separate adaptation location from capacity.

To test whether non-COCO difficulty is specific to our pipeline, we evaluate OVSeg, CAT-Seg, and MaskCLIP on ADE20K with the same COCO-overlap/non-overlap split: classes are COCO-overlap if their name matches a COCO panoptic category, and non-overlap otherwise.
\section{Empirical Analysis and Findings}
\label{empirical}

\subsection{Experimental Setup}
\label{empi:experimental}

\begin{figure*}[t]
\centering
\setlength{\tabcolsep}{4pt}
\renewcommand{\arraystretch}{1.0}

\begin{tabular}{ccccc}
\includegraphics[height=0.105\textwidth,keepaspectratio]{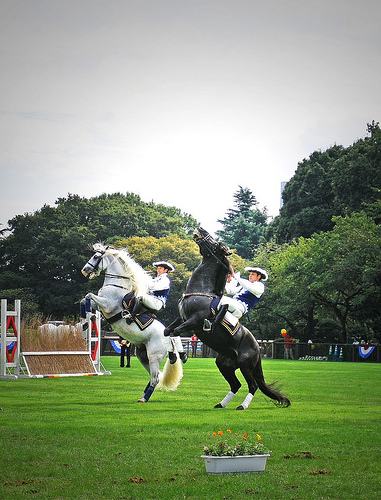} &
\includegraphics[height=0.105\textwidth,keepaspectratio]{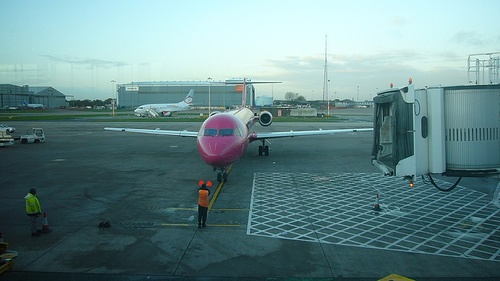} &
\includegraphics[height=0.105\textwidth,keepaspectratio]{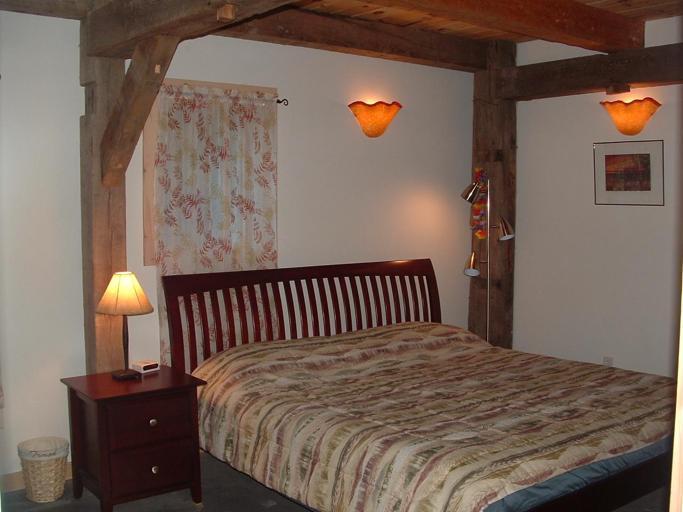} &
\includegraphics[height=0.105\textwidth,keepaspectratio]{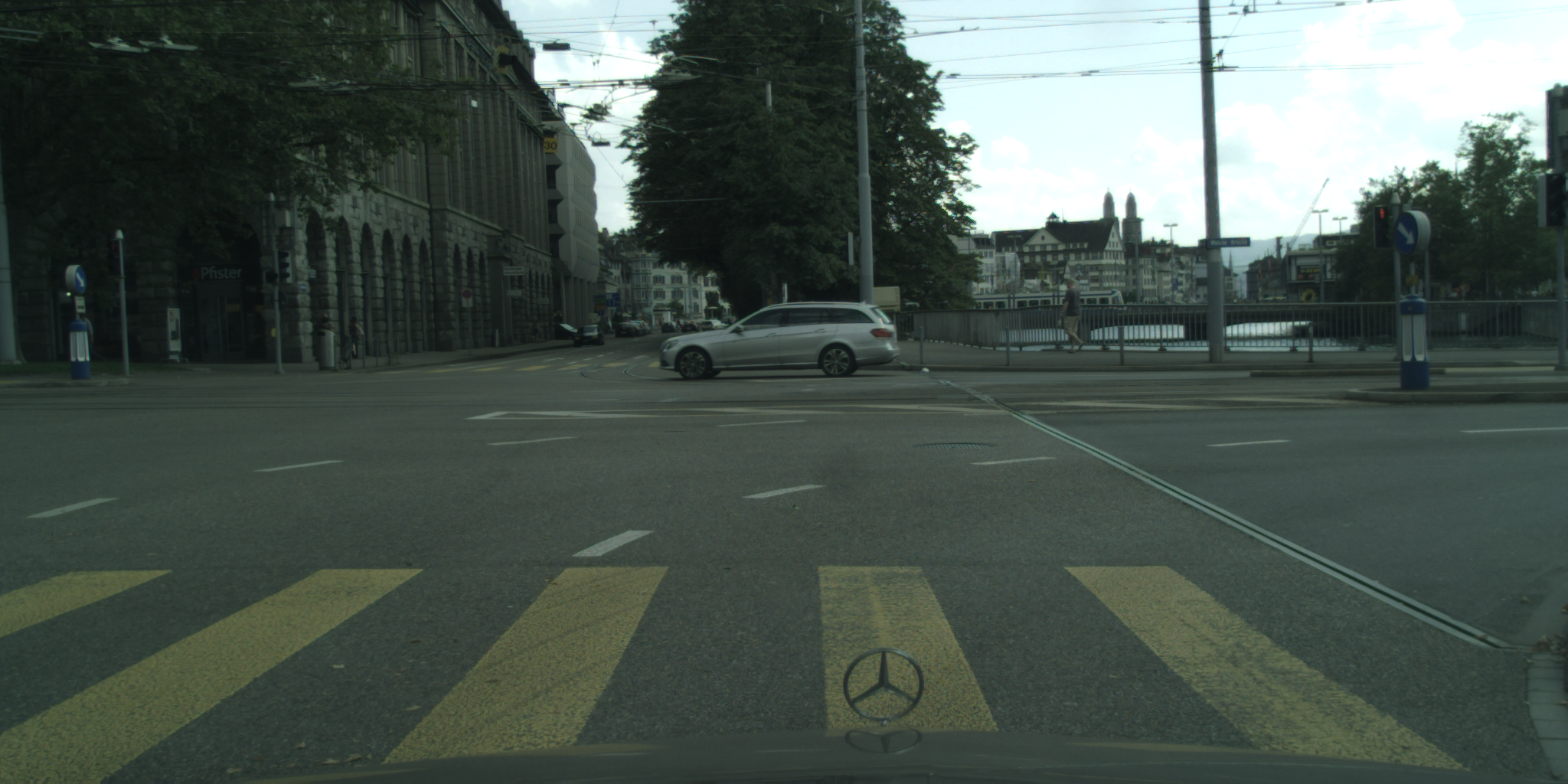} &
\includegraphics[height=0.105\textwidth,keepaspectratio]{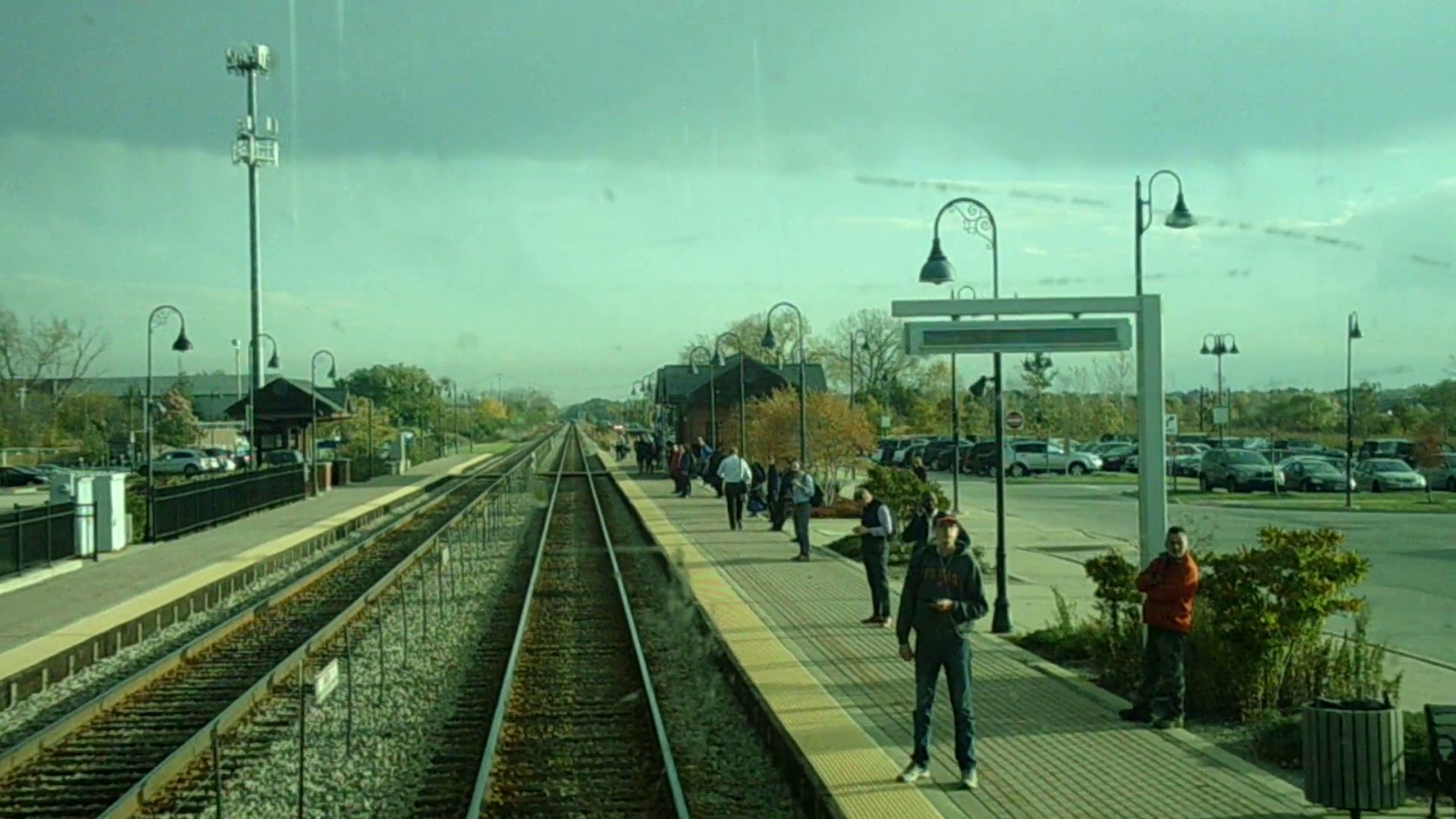} \\[-1mm]

\small (a) COCO & 
\small (b) Pascal & 
\small (c) ADE20K & 
\small (d) Cityscapes & 
\small (e) RailSem19
\end{tabular}

\caption{
\textbf{Evaluation domains.}
The benchmark suite spans in-domain (COCO), near-domain (Pascal), and cross-domain (ADE20K, Cityscapes, RailSem19).
}
\vspace{-1em}
\label{fig:dataset_grid}
\end{figure*}

All ODISE-based variants are trained only on COCO panoptic, including the text adapters and visual-side LoRA, and then evaluated under different domain regimes as shown in Fig.~\ref{fig:dataset_grid}. We train four LoRA variants (ranks 4, 8, 16, and 32) using the same training schedule (learning rate of 1e-4, a batch size of 8, and on 8 NVIDIA A100 40GB GPUs) to study sensitivity to the adaptation budget.

\begin{itemize}
    \item \textbf{In-domain:} COCO validation, reported with PQ and semantic mIoU.
    \item \textbf{Near-domain:} Pascal VOC 21 and Pascal Context 59, reported with semantic mIoU.
    \item \textbf{Cross-domain (OVPS):} ADE20K for scene parsing; Cityscapes, Mapillary Vistas, and WD2 for urban driving; and RailSem19 for railway scenes. Performance is measured using semantic mIoU for RailSem19, and panoptic metrics (PQ, SQ, RQ) for the others.
\end{itemize}

WD2 uses the 76-class validation split with 638 images, and RailSem19 uses 500 validation images with 19 semantic classes. For OVSeg, CAT-Seg, and MaskCLIP, we use the ADE20K-150 validation split (2000 images, 150 classes). The COCO-overlap split identifies 49--55 classes with a name matching a COCO panoptic category (the exact number varies slightly by method because rare classes that do not appear in the validation set are excluded from mIoU computation), the remaining 95--101 classes form the non-overlap set.


\subsection{The Cross-Domain Performance Inversion}
\label{empi:crossdomain}

We first test whether stronger dense visual features remain beneficial once they pass through a fixed OVPS recognition interface. Table~\ref{tab:main_benchmark} reveals a source-to-target sign reversal. \modelname{} improves COCO and near-domain metrics, so the substitution is partially working, however, it degrades sharply on urban and rail cross-domain targets. This split is the first indication that better dense features and better open-vocabulary transfer are not equivalent: the features can improve source-domain performance while failing to remain compatible with the downstream recognition interface under domain shift.

The backbone controls make the failure more specific. O-DINOv2 and O-ConvNeXt use the same ODISE-style decoder, CLIP recognition branch, and inference interface, yet they preserve urban performance. Thus, the collapse is not explained by replacing ODISE's diffusion extractor in general. It points instead to a DINOv3--ODISE compatibility issue: the new region features improve source- and near-domain performance but interact poorly with the fixed CLIP-based recognition and inference pipeline on shifted targets. ADE20K is the main exception to the urban/rail pattern; \modelname{} remains close to ODISE before adaptation, and we revisit this behavior in Sec.~\ref{empi:where}.

\begin{table*}[t]
\centering
\caption{Main benchmark results. \modelname{} improves source and near-domain metrics but collapses on urban/rail cross-domain targets, unlike the DINOv2 and CLIP-aligned ConvNeXt controls.}
{\footnotesize
\setlength{\tabcolsep}{4pt}
\begin{tabular}{l | c c | c c | c c c c c}
\toprule
\multirow{3}{*}{Model}
& \multicolumn{2}{c|}{\textbf{In-domain}}
& \multicolumn{2}{c|}{\textbf{Near-Domain}}
& \multicolumn{5}{c}{\textbf{Cross-Domain}} \\
& PQ & mIoU
& \multicolumn{2}{c|}{mIoU}
& \multicolumn{4}{c}{PQ}
& mIoU \\
& \multicolumn{2}{c|}{COCO} 
& VOC21 & Ctx59
& ADE & CityS. & Map. & WD
& RS19 \\
\midrule
ODISE
& 55.4 & 65.2
& 82.7 & 55.3
& 22.5 & 40.9 & 17.1 & 12.4
& \textbf{29.0} \\
FC-CLIP
& 54.4 & 63.7
& 81.8 & 58.4
& \textbf{26.8} & \textbf{44.0} & \textbf{18.2} & \textbf{13.0}
& 23.4 \\
\midrule
O-ConvNeXt
& 51.8 & 60.8
& 81.1 & 50.6
& 14.9 & 40.8 & 15.6 & 12.0
& \textbf{29.0} \\
O-DINOv2
& 54.8 & 64.3
& 84.0 & 52.5
& 15.8 & 41.8 & 15.6 & 12.2
& 26.7 \\
\modelname
& \textbf{57.0} & \textbf{67.4}
& \textbf{86.2} & \textbf{59.2}
& 21.5 & 19.2 & 7.1 & 3.6
& 8.0 \\
\bottomrule
\end{tabular}
}
\vspace{-1em}

\label{tab:main_benchmark}
\end{table*}

FC-CLIP provides an external reference rather than a controlled intervention. Its competitive cross-domain performance indicates that the target datasets are not intrinsically beyond mask-based OVPS. The remaining question is therefore where the \modelname{} pipeline loses these targets: at proposal generation, at inference retention, or at semantic assignment. We answer this next with the stage-wise decomposition.

\subsection{The Failure Is Recognition-Dominated, and Controls Confirm It}
\label{empi:failure}

We next localize where the cross-domain collapse occurs. Table~\ref{tab:pq_stagewise} reports stage-wise diagnostics for all five models across four cross-domain datasets.

\paragraph{The O-DINOv2 and O-ConvNeXt contrast.}
The controls show stage profiles that are nearly identical to ODISE on every dataset. On ADE20K, O-DINOv2 retains 45.4\% of matched proposals and labels 79.8\% correctly, and O-ConvNeXt retains 43.9\% and labels 77.9\% correctly. Both are close to ODISE's 46.0\% and 82.4\%. On Cityscapes, all three controls retain roughly 52\% of proposals with over 93\% label accuracy. This consistency across three different visual backbones in the same pipeline makes the O-DINOv3 collapse stand out more sharply. The failure does not emerge from replacing the diffusion extractor, nor from using a newer or stronger visual foundation. Among the evaluated backbone substitutions, the cross-domain collapse is specific to DINOv3.

\paragraph{The O-DINOv3 cascade.}
On Cityscapes, raw recall drops from ~67\% (ODISE) to 52.7\% (O-DINOv3), a meaningful but not catastrophic reduction in proposal availability. Retained recall drops from 53.0\% to 26.7\%, a collapse that far exceeds the raw recall gap. This means a large fraction of O-DINOv3's available proposals are thrown away by the inference pipeline. Label accuracy falls too (94.1\% to 81.1\%), so retained proposals are also harder to classify correctly. On Mapillary, SQ barely moves (50.5 vs.\ 51.2), but retained recall and label accuracy drop substantially, indicating that spatially plausible masks are being generated but not retained or recognized. WD2 is the most severe case: both SQ and RQ fall, suggesting that under larger appearance shifts the grouping quality degrades alongside the recognition failure.

\paragraph{What this rules out.}
The three-model control comparison allows a cleaner attribution than an O-DINOv3-vs-ODISE comparison alone. Because O-DINOv2 and O-ConvNeXt do not produce the cascade, the failure cannot be attributed to the Mask2Former~\cite{cheng2022masked} decoder, the CLIP recognition branch, or the panoptic inference rules alone. It is in how DINOv3 region features interact with those fixed components. This motivates us to look at the feature geometry directly, which we do next.

\begin{table*}[t]
\centering
\caption{
PQ decomposition and stage-wise diagnostics across all models and cross-domain benchmarks.
}
{\footnotesize
\setlength{\tabcolsep}{4pt}
\begin{tabular}{l l | ccc | ccc}
\toprule
\multirow{2}{*}{\textbf{Dataset}}
& \multirow{2}{*}{\textbf{Model}}
& \multicolumn{3}{c|}{\textbf{PQ Decomposition}}
& \multicolumn{3}{c}{\textbf{Stage-Wise Cascade}} \\
&
& \textbf{PQ} & \textbf{SQ} & \textbf{RQ}
& \textbf{Raw} & \textbf{Retained} & \textbf{Label} \\
& & & & & \textbf{Recall} & \textbf{Recall} & \textbf{Acc.} \\
\midrule
\multirow{5}{*}{ADE20K}
& ODISE     & 22.5 & 63.8 & 27.1 & 66.4 & 46.0 & 82.4 \\
& O-ConvNeXt & 14.9 & 55.7 & 18.5 & 63.5 & 43.9 & 77.9 \\
& O-DINOv2  & 15.8 & 56.4 & 19.2 & 64.6 & 45.4 & 79.8 \\
& \modelname & 21.5 & 56.6 & 26.0 & 60.7 & 28.0 & 66.6 \\
& FC-CLIP   & 26.8 & 71.2 & 32.3 & 72.6 & 46.0 & 83.8 \\
\midrule
\multirow{5}{*}{Cityscapes}
& ODISE     & 40.9 & 71.5 & 49.2 & 67.4 & 53.0 & 94.1 \\
& O-ConvNeXt & 40.8 & 70.2 & 49.8 & 61.3 & 52.1 & 93.8 \\
& O-DINOv2  & 41.8 & 70.8 & 50.7 & 62.9 & 52.7 & 93.6 \\
& \modelname & 19.2 & 62.9 & 23.6 & 52.7 & 26.7 & 81.1 \\
& FC-CLIP   & 44.0 & 75.4 & 53.6 & 70.1 & 46.1 & 96.6 \\
\midrule
\multirow{5}{*}{Mapillary}
& ODISE     & 17.1 & 51.2 & 21.2 & 28.6 & 19.0 & 90.2 \\
& O-ConvNeXt & 15.6 & 50.3 & 19.3 & 26.2 & 19.3 & 90.1 \\
& O-DINOv2  & 15.6 & 50.5 & 19.4 & 26.6 & 19.2 & 90.7 \\
& \modelname &  7.1 & 50.5 & 9.2 & 22.8 & 11.2 & 70.4 \\
& FC-CLIP   & 18.2 & 56.4 & 22.8 & 31.2 & 20.1 & 87.6 \\
\midrule
\multirow{5}{*}{WD2}
& ODISE     & 12.4 & 30.4 & 15.6 & 44.2 & 29.9 & 88.3 \\
& O-ConvNeXt & 12.0 & 30.2 & 14.8 & 41.1 & 29.7 & 88.0 \\
& O-DINOv2  & 12.2 & 30.3 & 15.0 & 42.2 & 30.3 & 88.9 \\
& \modelname &  3.6 & 19.8 & 4.6 & 36.0 & 15.9 & 54.2 \\
& FC-CLIP   & 13.0 & 42.0 & 16.3 & 46.8 & 21.2 & 79.5 \\
\bottomrule
\end{tabular}
}
\vspace{1mm}
\label{tab:pq_stagewise}
\end{table*}

\subsection{The Information Is Present, but the CLIP Readout Cannot Access It}
\label{empi:information}

\begin{table}[t]
\centering
\caption{ADE20K linear-probe diagnostic on matched proposal embeddings.}
{\footnotesize
\setlength{\tabcolsep}{4pt}
\begin{tabular}{l | c c | c c}
\toprule
\multirow{2}{*}{Model}
& \multicolumn{2}{c|}{All Matched Proposals}
& \multicolumn{2}{c}{Retained Proposals} \\
& CLIP Branch & Linear Probe & CLIP Branch & Linear Probe \\
\midrule
ODISE & 60.5 & 75.9 & 74.8 & 83.9 \\
\modelname{} & \textbf{28.8} & \textbf{57.3} & \textbf{49.9} & \textbf{75.3} \\
\bottomrule
\end{tabular}
}
\vspace{1mm}
\label{tab:linear_probe}
\end{table}

The stage-wise analysis shows where proposals are lost, but not whether retained or available regions are semantically readable by the CLIP recognition branch. We now test whether \modelname{} fails due to features lacking semantic content, or because the frozen CLIP text prototypes cannot extract that content from DINOv3's geometry (see 
Table~\ref{tab:linear_probe}). Under the frozen branch, \modelname{} achieves only 28.8\% top-1 accuracy on matched ADE20K proposals, less than half of ODISE's 60.5\%. A linear probe on the same embeddings reaches 57.3\% compared with 28.8\% for the frozen branch. Class-discriminative information is therefore present in the embeddings, but the frozen CLIP branch does not access it reliably.

We further measure region-text alignment on matched proposals: for each proposal with ground-truth class $y$, we record the top-1 accuracy, the correct-class cosine similarity, and the margin over the second-best prototype. On COCO-overlap proposals, \modelname{} has top-1 of 35.2\% and a negative margin ($-$0.079). O-DINOv2 and O-ConvNeXt reach 67.7\% and 64.8\% with positive margins ($+$0.104 and $+$0.081). Even for familiar COCO-adjacent classes, DINOv3 embeddings are misaligned with CLIP text prototypes, indicating that the failure is backbone-specific rather than a vocabulary difficulty. On non-COCO proposals, raw top-1 is low for \emph{all} systems: 7.6\%, 11.8\%, 12.2\%, and 13.7\% for O-DINOv3, O-ConvNeXt, O-DINOv2, and FC-CLIP. This shared non-overlap difficulty motivates Sec.~\ref{empi:vocabboundary}.

\subsection{Vocabulary-Boundary Difficulty Is Shared Across CLIP-Based Systems}
\label{empi:vocabboundary}
\begin{table}[t]
\centering
\caption{ADE20K-150 mIoU under the COCO-overlap/non-overlap split. OVPS PQ scores are not directly comparable to mIoU but confirm the same vocabulary-boundary pattern in our models.}
{\footnotesize
\setlength{\tabcolsep}{3pt} 
\begin{tabular}{l | c | c c | c c | l | c}
\toprule
\multirow{2}{*}{CLIP Method} & \multirow{2}{*}{mIoU} & \multicolumn{2}{c|}{COCO-overlap} & \multicolumn{2}{c|}{Non-overlap} & \multirow{2}{*}{OVPS Model} & \multirow{2}{*}{PQ} \\
& & $n$ & mIoU & $n$ & mIoU & & \\
\midrule
MaskCLIP (ViT-B/16)     & 11.4 & 55 & 16.3 & 95  & 8.5  & O-DINOv3      & 21.5 \\
OVSeg (Swin-B+ViT-L/14) & 29.6 & 49 & 43.5 & 101 & 22.8 & +LoRA & 16.0 \\
CAT-Seg (ViT-L/14@336)  & 37.8 & 55 & 50.4 & 95  & 30.5 & ODISE         & 22.5 \\
\bottomrule
\end{tabular}
}
\vspace{-1em}
\label{tab:vocab_boundary}
\end{table}

Raw top-1 accuracy on non-COCO proposals is low throughout---7.6\%, 11.8\%, 12.2\%, and 13.7\% for \modelname{}, O-ConvNeXt, O-DINOv2, and FC-CLIP, respectively. Is this non-overlap difficulty specific to our pipeline, or structural to CLIP-based recognition? To answer this, we evaluate MaskCLIP~\cite{tu2023open}, OVSeg~\cite{liang2023ovseg}, and CAT-Seg~\cite{cho2024catseg} on ADE20K-150 with the same overlap/non-overlap split. 

Table~\ref{tab:vocab_boundary} shows the pattern is shared: MaskCLIP has a 7.8-point gap (16.3 vs 8.5), OVSeg 20.7 points (43.5 vs 22.8), and CAT-Seg 19.9 points (50.4 vs 30.5). All three show a large, consistent non-overlap deficit despite spanning very different architectures and training strategies. The vocabulary-boundary limitation therefore extends beyond our specific backbone and pipeline and is shared across the evaluated CLIP-based recognition systems. This explains why neither text-side nor visual-side adaptation fully recovers ADE20K: the evaluated CLIP-based recognition systems perform substantially worse on non-overlap categories.

\subsection{Where Does the Alignment Burden Live?}
\label{empi:where}
The previous diagnostics show that \modelname{} often produces usable region evidence, but the fixed CLIP recognition interface does not retain or classify it reliably. But where should this recoverable gap be repaired? Table~\ref{tab:adapter_interventions} compares five adaptation sites within the same \modelname{} pipeline: the text embeddings, the visual backbone, the retention rule, a capacity-matched large text adapter (LTA), and visual-side LoRA combined with retention calibration (LoRA+CMR).

The interventions separate three different failure sources. \textit{Text-side adaptation} substantially recovers urban and rail performance, showing that recalibrating the recognition space can repair many region-text errors. However, it degrades ADE20K, indicating that moving the text embeddings using COCO supervision is sensitive to vocabulary overlap. This supports the vocabulary-boundary result (Sec.~\ref{empi:vocabboundary}): source-anchored text calibration helps COCO-adjacent targets but does not transfer uniformly to broader scene-parsing categories. The small text adapter already achieves most text-side recovery; scaling to LTA adds only marginal and inconsistent improvements. \textit{Visual-side LoRA} gives the most consistent recovery on the urban and rail targets. It is trained only on COCO, keeps the base DINOv3 backbone frozen, and adds only 2.7M trainable parameters, yet it exceeds ODISE on Cityscapes, WD2, and RailSem19 and nearly recovers Mapillary. 
This is 
evidence that the dominant recoverable failure is not missing semantic information, but a visual-to-CLIP interface gap: adjusting the region features before recognition is more reliable than moving the text space after the fact. \textit{Retention-only CMR} improves Cityscapes (where the stage-wise analysis showed heavy suppression) but has little effect elsewhere and degrades ADE20K, suggesting that retention is not the primary bottleneck when proposals are present but geometrically misaligned. Combining LoRA with CMR does not improve over LoRA alone, indicating that retention errors become subordinate once visual features are adapted. 

\begin{table}[t]
\centering
\caption{
Alignment and retention interventions for \modelname{}. 
}
{\footnotesize
\setlength{\tabcolsep}{4pt}
\begin{tabular}{l l r | c c c c c}
\toprule
Intervention & Locus & Params
& ADE & CityS. & Map. & WD & RS19 \\
\midrule
No adapter
& -- & 24.7M
& 21.5 & 19.2 & 7.1 & 3.6 & 8.0 \\
\midrule
 + Text adapter (small)
& text-side & +0.10M
& 15.8 & 42.2 & 14.7 & 12.6 & 24.8 \\
 + LTA
& text-side & +2.62M
& 15.7 & 44.0 & 15.7 & 13.2 & 27.1 \\
\midrule
 + LoRA r16
& visual-side & +2.7M
& 16.0 & \textbf{45.6} & \textbf{16.1} & \textbf{14.5} & \textbf{29.4} \\
 + LoRA r16 + CMR
& visual+ret. & +2.7M
& 15.6 & 45.1 & 15.6 & 14.3 & 29.4 \\
\midrule
 + CMR
& retention-only & +0
& 9.9 & 23.4 & 8.2 & 3.3 & 8.0 \\
\midrule
ODISE (reference)
& -- & 28.2M
& \textbf{22.5} & 40.9 & 17.1 & 12.4 & 29.0 \\
\bottomrule
\end{tabular}}
\label{tab:adapter_interventions}
\end{table}

\textbf{Rank sensitivity.} Table~\ref{tab:lora_rank} sweeps LoRA ranks from 4 to 32. Rank 4 already recovers most of the urban/rail gap at one quarter the parameter count of rank 16. Performance peaks at rank 16, and rank 32 shows no further improvement and regresses slightly on Mapillary and RailSem19, suggesting that a relatively low-dimensional adaptation is sufficient to recover much of the interface gap. ADE remains flat across all ranks, indicating limited sensitivity to visual adaptation capacity over the evaluated rank range.

\begin{table}[t]
\centering
\caption{\modelname{} LoRA rank sweep. All trained 92k iterations on COCO panoptic}
{\footnotesize
\setlength{\tabcolsep}{3.5pt}
\begin{tabular}{l r | c c c c c}
\toprule
LoRA rank & Params & ADE & CityS. & Map. & WD & RS19 \\
\midrule
No adapter & -- & \textbf{21.5} & 19.2 & 7.1 & 3.6 & 8.0 \\
\midrule
r4  & +0.67M & 16.1 & 43.4 & \textbf{16.1} & 13.8 & 29.0 \\
r8  & +1.34M & 16.5 & 42.6 & 16.0 & 13.4 & 27.2 \\
r16 & +2.7M  & 16.0 & \textbf{45.6} & \textbf{16.1} & \textbf{14.5} & \textbf{29.4} \\
r32 & +5.4M  & 16.1 & 43.8 & 15.4 & 13.3 & 28.6 \\
\bottomrule
\end{tabular}
}
\vspace{-3mm}
\label{tab:lora_rank}
\end{table}

\textbf{ADE20K anomaly.} 
Unlike driving/rail targets that share COCO's taxonomy, two-thirds of ADE20K's 150 classes are non-COCO. Unadapted \modelname{} holds 21.5 PQ because ADE20K's natural scenes visually resemble COCO, avoiding severe suppression. However, training LTA or LoRA solely on COCO induces source-taxonomy overfitting: warping the projection manifold toward the 133 COCO prototypes sharpens in-vocabulary boundaries but distorts out-of-vocabulary geometry, collapsing non-overlap predictions (21.5 $\rightarrow$ 16.0 PQ). Conversely, ODISE retains 22.5 PQ because its diffusion features are pre-aligned with broad open-domain text prompts during pretraining.
\section{Qualitative Analysis}
\label{qualitative}

\begin{figure*}[t]
\centering
\def\svgwidth{0.95\textwidth}
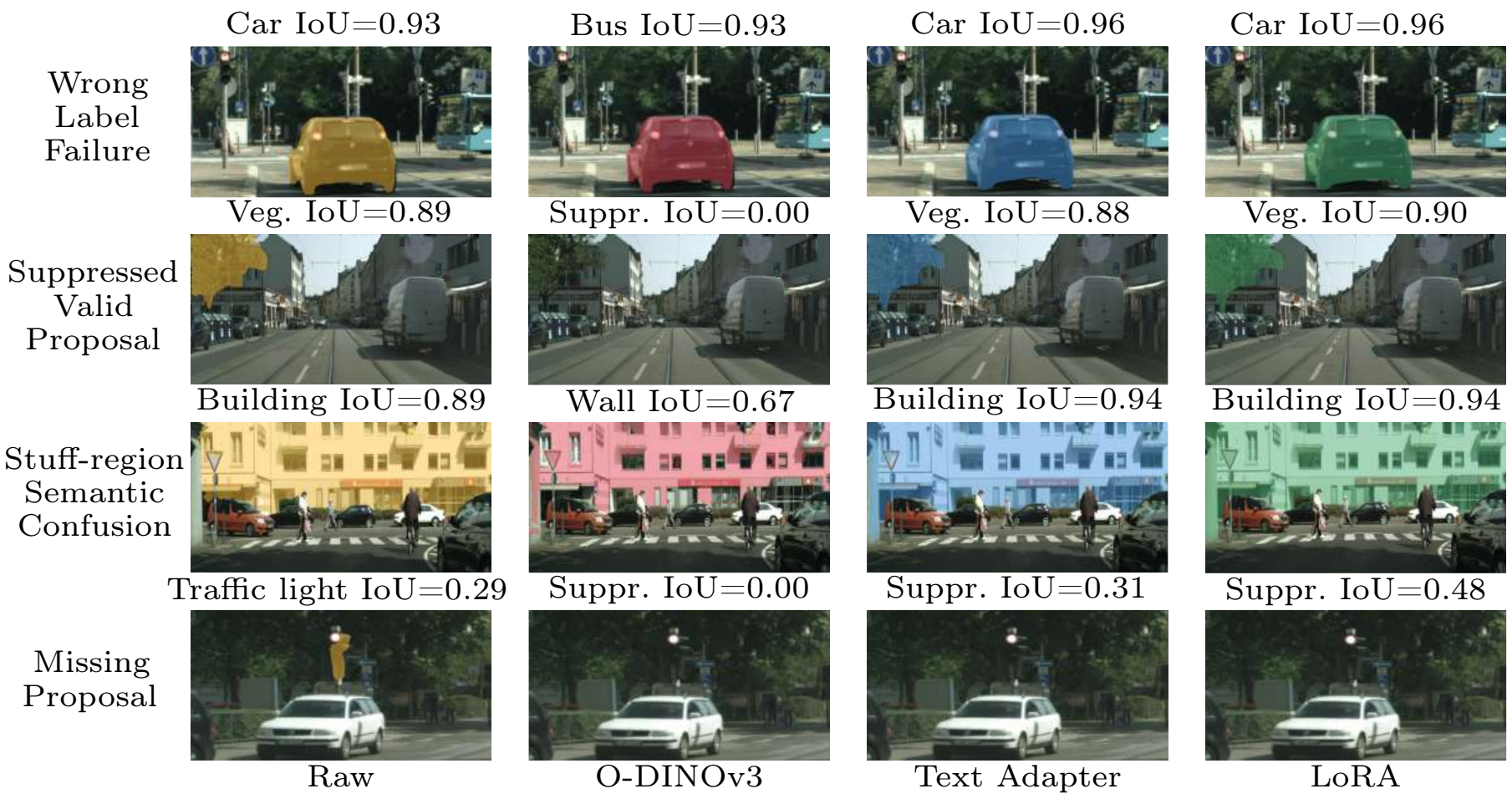
\caption{
\textbf{Failure taxonomy for \modelname{} on Cityscapes.}
Each row illustrates one stage-wise failure mode. 
}
\label{fig:failure_taxonomy}
\vspace{-5mm}
\end{figure*}
The stage-wise cascade produces qualitatively distinct failure patterns. When valid proposals exist but are suppressed or mislabeled, both text-side and visual-side adaptation can restore them. Instead, LoRA gives more consistent recovery. When the raw proposal itself is absent or below the matching threshold, recognition-side changes cannot recover the missing proposal. Figure~\ref{fig:failure_taxonomy} illustrates these modes on Cityscapes.

\begin{figure*}[t]
\centering
\def\svgwidth{0.95\textwidth}
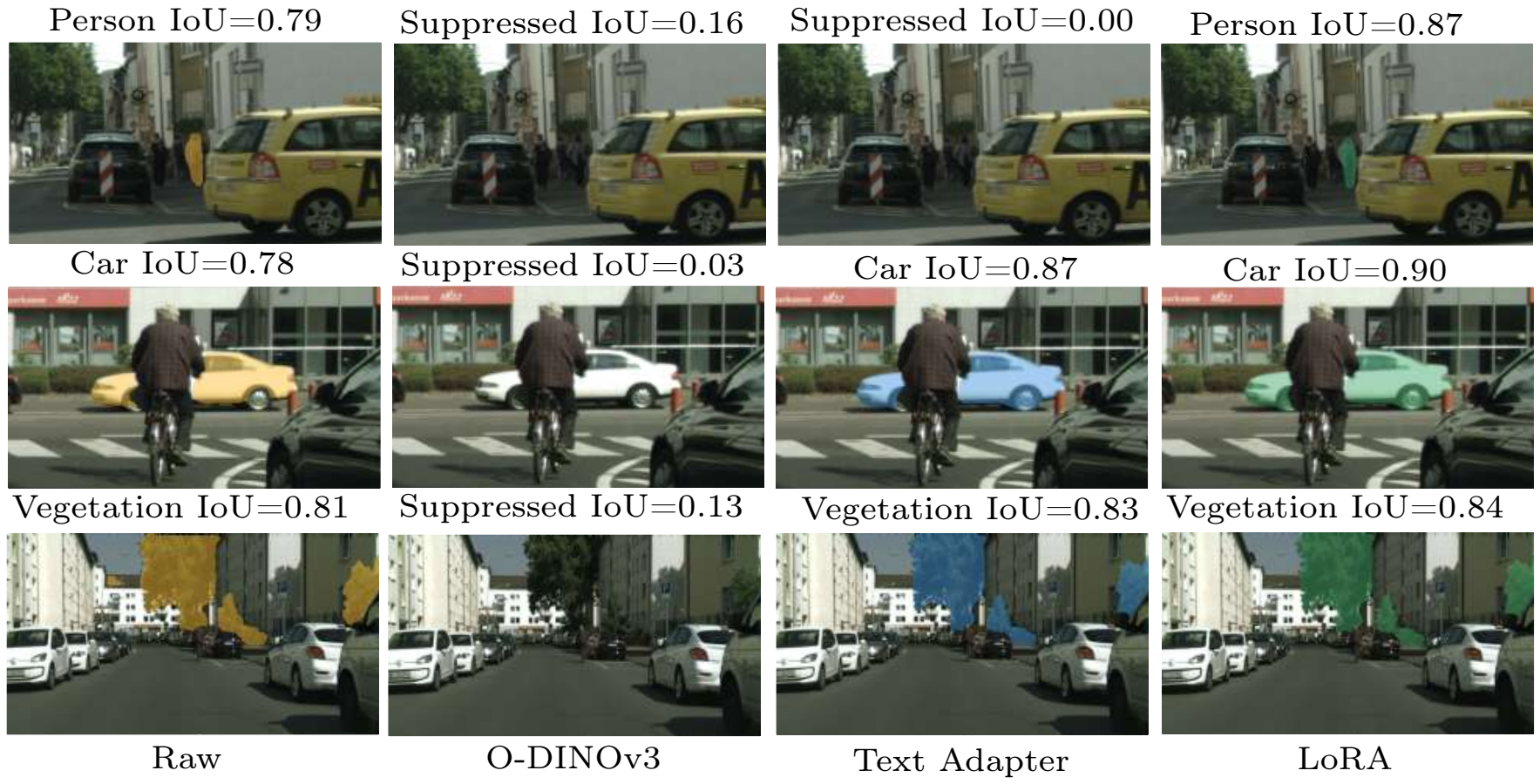
\caption{
\textbf{Qualitative comparison of adaptation effects.}
Each row shows a valid-proposal failure and the corresponding predictions from \modelname{}, text-side adaptation, and visual-side LoRA. 
}
\label{fig:repair_comparison}
\vspace{-5mm}
\end{figure*}

Figure~\ref{fig:repair_comparison} visualizes where the recoverable part of the gap is repaired. When a useful raw proposal is available, changing the recognition interface can restore the correct region. The text adapter is sufficient for some COCO-adjacent categories such as car and vegetation, but it is not uniformly reliable, as shown by the person example. Visual-side LoRA recovers all three examples, matching the quantitative result that adapting the region features before CLIP-based recognition gives the most consistent urban/rail recovery.

Together, the qualitative examples make the main empirical message concrete. \modelname{} often has usable region evidence, but the fixed recognition and inference interface can suppress or misread it. Adaptation can help when a usable candidate exists, and the failure occurs at the interface. However, it cannot fully repair errors where the proposal itself is missing or below the matching threshold. ADE20K examples in the supplement further illustrate the remaining difficulty on non-overlap categories.

\section{Discussion}
\label{discussion}

Our results identify two separable failure axes in cross-domain OVPS. First, a \textbf{visual-to-CLIP interface gap}: stage-wise cascades, alignment margins, and linear probing all show that DINOv3 region features retain class-discriminative semantics, but require visual-side LoRA to align their native geometry with the frozen CLIP readout. The second is a \textbf{vocabulary-boundary limitation}: ADE20K non-COCO categories are poorly covered by the CLIP recognition space across the evaluated systems. Table~\ref{tab:vocab_boundary} shows the same overlap/non-overlap gap in OVSeg, CAT-Seg, and MaskCLIP---methods spanning very different architectures and training scales. The ADE20K failure extends beyond DINOv3 and is shared across the evaluated CLIP-based recognition systems. Visual-side adaptation trained solely on a fixed source taxonomy cannot resolve this deficit on non-overlap categories. Resolving this non-overlap deficit requires moving beyond fixed source supervision, such as via training-free retrieval prototypes (e.g., ReME~\cite{xuan2025reme}), VLM-generated pseudo-captions, or LLM-enriched class descriptions.

The capacity-matched comparison between LTA (+2.62M, text-side) and LoRA r16 (+2.7M, visual-side) is the cleanest test of this distinction: at equal parameter count, visual-side adaptation wins on all urban and rail targets. The bottleneck is not text adapter capacity but adaptation placement. Adding CMR on top of LoRA provides little additional gain, suggesting that once the representation is aligned, retention suppression is no longer the active bottleneck.

The stage-wise framing (raw recall, retained recall, label accuracy) is more diagnostic than a single PQ number: missing proposals, suppressed valid masks, and wrong-label errors require different remedies. We hope this protocol is useful beyond our specific study.

\textbf{Limitations.} Our intervention is centered on ODISE; FC-CLIP is an external reference only. We evaluate one DINOv3 variant (ViT-H/16+) and one training configuration. The MaskCLIP evaluation uses single class names rather than all ADE20K synonyms, contributing to the gap from the published number, though the overlap/non-overlap ratio remains meaningful.

\section{Conclusion}
\label{conclusion}

We show that stronger dense vision foundation backbones are not plug-and-play upgrades for cross-domain OVPS. In a controlled ODISE intervention, DINOv3 improves source-domain results but collapses on urban and rail targets, while DINOv2 and CLIP-aligned ConvNeXt controls do not. Stage-wise diagnostics, region-text alignment, and linear probing reveal two separable failure axes: (1) a geometric mismatch between DINOv3 region features and the frozen CLIP recognition branch, largely repairable by visual-side LoRA; and (2) a vocabulary-boundary limitation on ADE20K non-COCO categories that is shared by OVSeg, CAT-Seg, and MaskCLIP, indicating that this limitation extends beyond DINOv3 and our specific pipeline. Visual-side LoRA (+2.7M) outperforms a capacity-matched large text adapter (+2.62M) on all urban and rail targets, indicating that adaptation placement matters beyond parameter count. These results argue for interface-aware backbone adaptation and stage-wise diagnostics as standard practice in cross-domain OVPS evaluation.

\section*{Acknowledgements}
The computational results presented have been achieved using the Vienna Scientific Cluster (VSC). This work was supported by the Austrian Research Promotion Agency (FFG) under project number FO999930983.

%
%
\bibliographystyle{splncs04}
\bibliography{main}
\end{document}